*Monitoring spatial sustainable development: Semi-automated analysis of satellite and aerial images for energy transition and sustainability indicators.*

Eurostat Grant Agreement: 08143.2017.001-2017.408

# OVERVIEW OF THE AVAILABLE DATASETS

Date of issue: SEPTEMBER 18

|  | Name |
|---|---|
| Lead Author | Curier, R.L ( CBS) |
| Contributing Author | De Jong, T.J.A. (CBS)<br>Strauch, Katharina (IT.NRW)<br>Cramer, Katharina (IT.NRW)<br>Rosenski, Natalie (DESTATIS)<br>Schartner, Clara (DESTATIS)<br>Debusschere M. (Statbel)<br>Ziemons, Hannah (Universiteit Maastricht)<br>Iren, Deniz (Open Universiteit)<br>Bromuri, Stefano (Open Universiteit) |
| Reviewed by |  |
| Distribution List |  |

| Document change records | | | |
|---|---|---|---|
| Date | Issue | Affected | Description |
| 15-08-2018 | Draft | All | First Issue |
| 28-08-2018 | Draft | All | First draft |
| 10-09-2018 | 1.0 | All | Final first version – Draft update after all partner feedbacks |
| 25-09-2018 | 1.1 | All | Editing |

**Summary**

Solar panels are installed by a large and growing number of households due to the convenience of having cheap and renewable energy to power house appliances [JKK17]. Governments have been incentivising the use of this technology to foster a transition from an economy still relying on carbon based fuels, to a green technology with a low carbon footprint. As kilowatt prices from solar energy are continuously falling and the contribution of solar energy to overall energy production in the European Union (EU) is rapidly growing, business interest grows and so does governmental necessity to monitor and regulate renewable energy. This triggers interest in data on solar energy production and the distribution of solar panels.

In contrast to other energy sources solar installations are distributed very decentralized and spread over hundred-thousands of locations. On a global level more than 25% of solar photovoltaic (PV) installations were decentralized [REN21]. In Germany even about 40% of the total installed PV capacity was owned by private households in 2010. The effect of the quick energy transition from a carbon based economy to a green economy is though still very difficult to quantify. As a matter of fact the quick adoption of solar panels by households is difficult to track, with local registries that miss a large number of the newly built solar panels. This makes the task of assessing the impact of renewable energies an impossible task. Although models of the output of a region exist, they are often black box estimations.

This project´s aim is twofold: First automate the process to extract the location of solar panels from aerial or satellite images and second, produce a map of solar panels along with statistics on the number of solar panels. Further, this project takes place in a wider framework which investigates how official statistics can benefit from new digital data sources.

At project completion, a method for detecting solar panels from aerial images via machine learning will be developed and the methodology initially developed for BE, DE and NL will be standardized for application to other EU countries. In practice, machine learning techniques are used to identify solar panels in satellite and aerial images for the province of Limburg (NL), Flanders (BE) and North Rhine-Westphalia (DE). Besides, the project will also make use of existing registers such as information on the VAT returns by people buying solar panels and information acquired from the energy providers. As these registers are not complete, a main part of the project will therefore consist of using the machine learning approaches to fill the gap in the information in the register.

This document provides the reader with an overview of the various datasets which will be used throughout the project. The collection of satellite and aerial images as well as auxiliary information such as the location of buildings and roofs which is required to train, test and validate the machine learning algorithm that is being developed.

**Graphical Summary**

| Region of Interest | Resolution | Description |
|---|---|---|
| NRW (DE) | DOP20 20x20 cm<br>DOP10 10x10 cm | open source<br>open source summer 2018 |
| Limburg(NL) | LR 25x25 cm<br>HR 10x10 cm | open source |
| Flanders(BE) | 25x25 cm | open source |
| **Satellites Images** | | |
| Small area NRW | 10 m Sentinel<br>5 m Rapid Eye<br>1 m to 2 m, Spot 6 and7<br>Below 1 m Pleiades | |
| Limburg | 10m Sentinel<br>5m Rapid Eye<br>80cm TripleSat | Open source<br>NSO portaal free data for NL. |

# Contents

# Introduction

## 1.1. Context

In Europe, ongoing trends require urban energy systems that emit less carbon and use less energy as the current aim is to replace at least 30% of the demand for fossil fuels by renewable resources by 2050. To this end, big investments are made to develop smart renewable energy systems at all levels and current regulations focus on incremental measures to reduce energy consumption and greenhouse gas emissions. In North Western Europe the energy markets are moving rapidly and in recent years various studies have been undertaken to predict energy use and monitor the energy transition from fossil fuel to renewable energy sources such as solar, wind, biogas and geothermic etc. In view of these emission reduction targets, municipalities need to further their energy transition as they are the largest consumers of energy and emitters of greenhouse gas emissions through heating, air conditioning, manufacturing and transport. This project aims to contribute to a better understanding of the energy transition in terms of the use of photovoltaic solar panels at a regional to local level.

Current statistics on solar energy are based on a survey data from the importation of solar panels. The overall solar production is based on an estimate of installed capacity (charged panels) and an extensive number of production capacity per unit. In the Netherlands, the current methodology has an estimated uncertainty of 20% and solely provides national figures on a yearly basis while the energy transition creating a demand for information at regional level with shorter time scales. It is therefore of prior importance to provide a complete and detailed overview of the current solar panels installation.

## 1.2. Purpose and Scope

This project aims to contribute to a better understanding of the energy transition which is happening at regional to municipality level and is strongly related to the Sustainable Development Goals' s 7 and 11. We propose to analyse by means of artificial intelligence the information content from high resolution satellite and aerial images for the automatic detection and classification of solar panels. In practice, the process of extracting the location of solar panels from images will be automated to produce maps of solar panels spatial distribution along with regional statistics.

Besides the project will also make use of existing registers such as information on the VAT returns by people buying solar panels and information acquired from the energy providers. As these registers are not complete, a main part of the project will therefore consist of using the machine learning approaches to enrich the information in the register.

This report provides an overview of the available images and auxiliary information which will be used during this project. Available datasets for aerial images and administrative register are presented in Section 3, 4 and 5, for Netherlands, Germany and Belgium respectively.

Recently, a general consensus has been reached in Europe that satellite remote sensing is a viable means to provide a measurement-based characterization on a regional to a global scale as, they are in principle derived using a consistent approach over time and space. Nowadays, high resolution satellite and aerial images provide a wealth of regional statistical information on a regular basis. Hence, information such as building characteristics, land cover and land use can be determined from these images and if the resolution is high enough, even smaller objects, like solar panels can be detected. To this end, the added-value of various earth observation datasets is also evaluated, Section 6 provides with an overview of available datasets. In Section 7, a description of the software developed for the automatic download and processing of the aerial images is provided; Section 8 concludes this report and draws the lines for future explorations.

## 2. Related Work

Despite the fact that natural images as in the well-known ImageNet challenge are the benchmark for many image recognition deep learning models today, deep learning for image classification and object detection has also been applied to numerous use cases involving aerial or satellite images as for scene or land use classification [CPSV15, HXHZ15, LKAL16], road detection [FKG13, MFW16, MH10], vehicle detection [ALSL16, CXLP14], people detection [DOW16], invasive weed localization [HXS14] and buildings detection [PNdS15]. In particular, the contribution in [PNdS15] showed in an extensive experiment that CNN models perform better than most other low- and mid-level descriptors for image classification on aerial image and remote sensing data sets. In [BGM+15] then different deep learning models were compared for the task of classification and a Deep Belief Network (DBN) turned out to be the best performing algorithms. Since then several other examples indicated that even models which were pre-trained on other image types as natural images, generalize well to aerial photos, which demonstrates the generalization power of CNNs [MLD+16, FISA16]. Another example is the success of a pre-trained deep learning model on the UC Merced Land Use benchmark, which achieved an accuracy of 92.4% outperforming the other attempted models. The objects classified there were however considerably larger than solar panels, while the resolution was only slightly worse than at the Deep Solaris project.

The detection of solar panels comes with some additional challenges and has been investigated relatively rarely in the literature. In some early attempts to automatically infer the information of panel location from aerial images, traditional feature descriptors were used as an input to machine learning classifiers and already delivered usable results [MHC+15].

The creation of a large data set including over 19,000 solar panels annotated with the geospatial coordinates and border vertices in 2016 spurred research efforts. The images are taken over four cities in California with a minimum resolution of 30 cm per pixel [BSLJ+16]. A team from Duke university built several models for the task of classification and detection of solar panels, where CNNs demonstrated to produce the best results [MCB17]. Exploiting the advantages of such a large data set, the team also built its own VGG-inspired model which yielded a precision of roughly 0.95 at a recall rate of 0.8. Transfer learning however turned out not to be advantageous in their case [MCB17]. Similarly, a SegNet-inspired model, which is originally used for segmentation, but can be repurposed for object detection, could not deliver successful results [CWC+18].

Two other papers both reach accuracy levels of over 80% when applying CNNs to solar panel detection by creating their own labelled data sets and using relatively small self-made models.

On one hand, the current state of the art is interesting as this can be used to validate the results of the current study. On the other hand, the variety of formats and models, makes it difficult to compare the success of the existing approaches with respect to the data that will be collected.

This report hence aims to shed more light on the available sources of data in the European region comprising Limburg, Flanders and North Rhine Westphalia. Noticing that CNNs already proved fit for these tasks, the work done in this project will be used to investigate how different architectures, data pre-processing techniques and image resolutions affect the ability of the model to detect solar panels in the image.

# 3. Netherlands – Limburg

## 3.1. Aerial Images

Statistics Netherlands provides several years of aerial images for the whole of the Netherlands. The aerial images are updated every year with a new campaign. This means that per year the data up until the previous year is available. The aerial images have a resolution of 25cm ground pixel and are available as digital image files within the Statistics Netherlands. There are two digital image files for every year: one with RGB and one with IR data. In addition, the aerial images are also available as open data web services (WMS/WMTS) via http://www.pdok.nl, a joined effort of several Dutch authorities to provide reliable geodata. Though, these web services have a fair use policy that only allows around 10000 requests per month. The availability of several years of data will allow us to research generalization of machine learning models as well as development of solar panel availability in time. An example of the available data is given in Figure 1.

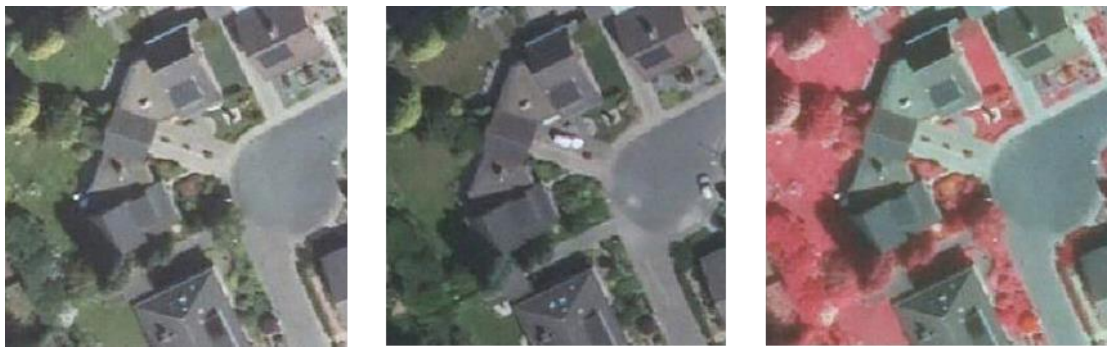

**Figure 1: Example DOP25 RGB images for 2016 (left panel), 2017 (middle panel) and IR for 2016 (right panel) available for The Netherlands.**

## 3.2. Administrative Datasets

### 3.2.1. Solar Panel Data

The location of more than 500000 solar panel installations are available as register data within Statistics Netherlands; both the postcode, house number and the building object id are present. In addition to that, there is information about the date the panels were installed. The geo location of each solar panel in the register is obtained by linking the solar panel address data with the geo location of each address. The register data is based on information retrieved from the energy providers on the one hand, VAT tax returns on the other. The register data about the solar panels is not complete, however. The register data therefore will be mainly used to create a training set for the machine learning models. In addition, it will also partly be used for validation of the machine learning results, but as the data is not complete, part of the validation will have to be done manually.

### 3.2.2. Building Polygons

All addresses and all the polygons of the buildings in the Netherlands are provided by the Dutch Cadastre. In addition, they are provided as an open data web service at the aforementioned http://www.pdok.nl and as a download at http://www.nlextract.nl. An example of these addresses and polygons can be seen in Figure 2

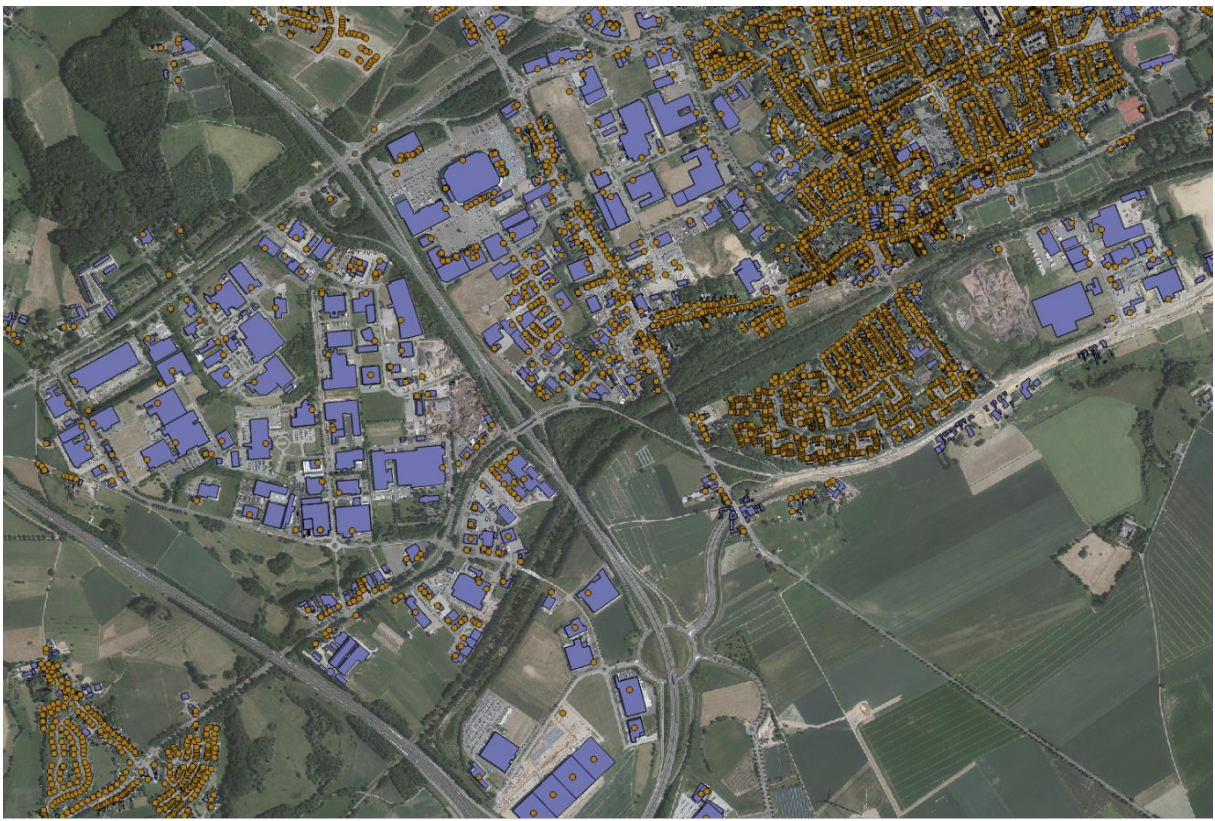

Figure 2: The house polygons and addresses provided by the Dutch Cadaster

Each building polygon also provides information about the status of the building, the purpose and the year it was built. The building polygons will be used for several purposes, like for example:

1. to generate a mask image that can be used to mask the part of the aerial images that are buildings,
2. filtering the cut-out areas of the aerial images to the areas that have buildings,
3. providing an address to detected solar panels; the building polygon will serve as a bounding box.

# 4. Germany – North Rhine-Westphalia

## 4.1. Aerial Images[1]

The district government in Cologne provides spatial-based data in a comprehensive infrastructure for North Rhine-Westphalia (NRW). Their product range contains high-resolution aerial images which combine the visual attributes of an aerial photograph with the spatial accuracy and reliability of a planimetric map[2]. These so-called digital orthophotos (DOP) cover the whole area of NRW and are updated every three years. The images are used, for example, for the planning process regarding road construction or environmental protection, soil science, archaeological investigations etc.

The data is provided as open data. The ground resolution of the aerial images was up to June 2018 20 cm (DOP20) and is now 10 cm (DOP10). By now only the DOP10 are available online (those are from 2015-2018)[3]. The user can download the data in tiles of 5000 pixel x 5000 pixel, which cover an area of 1 square kilometer, or in packages for each municipality.

Since the resolution is high, it is assumed that even small solar panels are visible in the images. Moreover the different resolutions DOP20 and DOP10 allow testing the algorithm on these different resolutions.

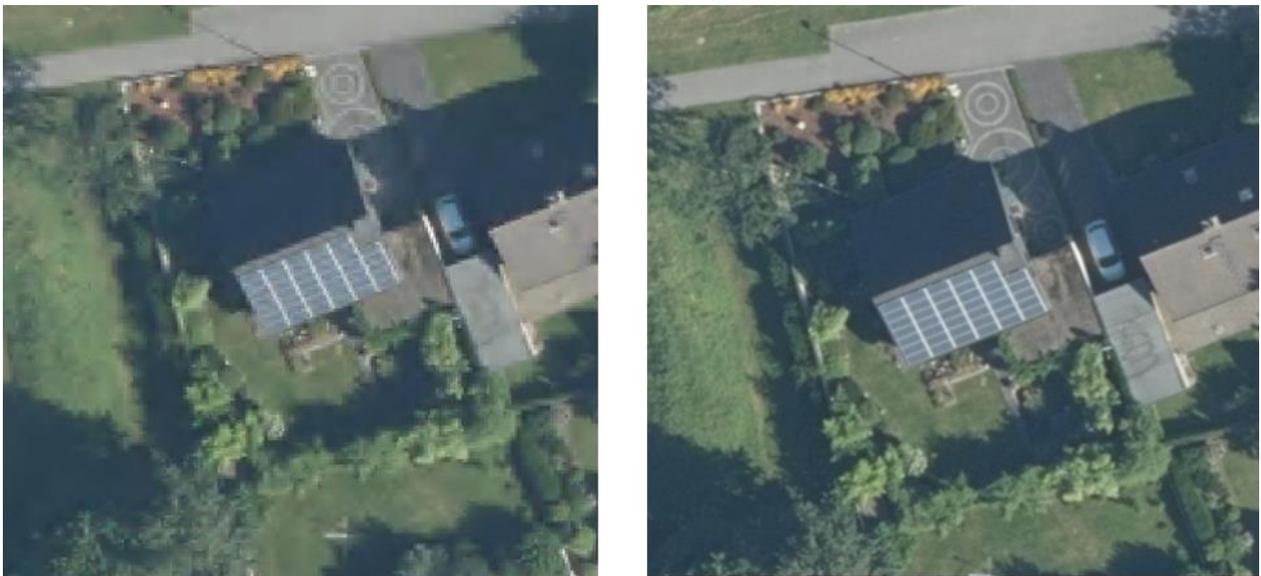

**Figure 3: Example of aerial images available in Germany at resolution DOP 20 (left) and DOP 10(right)Geobasisdaten der Kommunen und des Landes NRW © Geobasis NRW. No official standard version**

---

[1] The information regarding the aerial images is provided by the NRW district government in Cologne (https://www.bezreg-koeln.nrw.de/brk_internet/geobasis/luftbilderzeugnisse/digitale_orthophotos/index.html)

[2] Planimetric maps (also known as line maps) are two-dimensional maps which are only showing the horizontal positon of the natural and cultural features on the Earth's surface.

[3] The flight program for the photographs ("Bildflugprogramm") shows the flight plan for the collection of the images: https://www.bezreg-koeln.nrw.de/brk_internet/geobasis/luftbilderzeugnisse/digitale_luftbilder/bildflugprogramm.pdf
 https://www.tim-online.nrw.de/tim-online2/uebersicht.html?thema=dop.

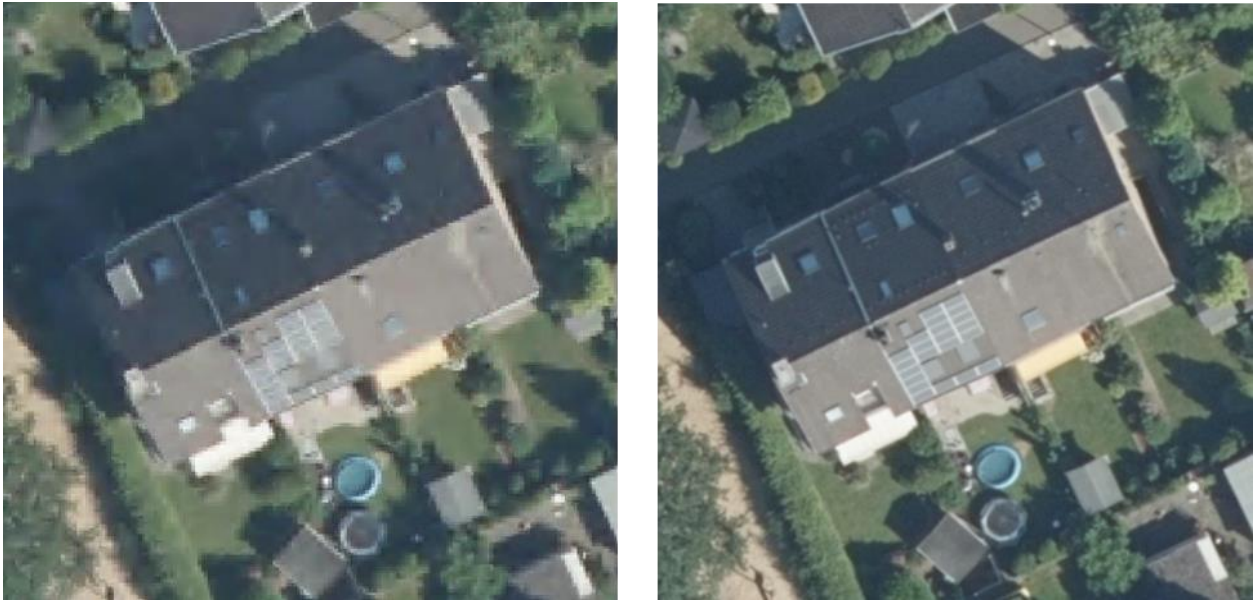

Figure 4: Example of aerial images available in Germany at resolution DOP 20 (left) and DOP 10(right).

## 4.2. Administrative Datasets

### 4.2.1. Solar Panel Data[4]

The Agency for Environment NRW (Landesamt für Natur- Umwelt- und Verbraucherschutz -- LANUV) provides data on different energy sources for NRW. One of those sources is renewable energy and more specific solar energy. The "Energieatlas NRW" shows the exact location of all photovoltaic panels with an electronic performance of more than 30kW. Smaller systems are located at the postcode level. The dataset includes the solar panels on rooftops and those which are located in open areas. The original data is provided by the federal network agency (Bundesnetzagentur= BNA) and transmission system operators. In contrast to the DOP20 and DOP10, the solar panel data is not available in an open data format.

To use the data from the "Energieatlas NRW" and to get moreover the exact location of systems with less performance than 30kW, IT.NRW and the LANUV agreed on specific terms of use. Due to those user terms IT.NRW was able to download the data via download link. Also the other project partners can download the data, after they sign the terms of use.

The dataset includes all solar panels which have been reported up to February 2018 and started to operate at the latest on December 31st 2017. Following information is included:

- Location
- Start of operations
- Performance
- Type (rooftop / open area)

---

[4] The information regarding the solar panel data is provided by the Landesamt für Natur- Umwelt- und Verbraucherschutz NRW (LANUV) (http://www.energieatlas.nrw.de/site/bestandskarte).

The data can be used to validate the findings of the algorithm. It must be noted that it is not allowed to publish the exact locations of the solar panels from the dataset.

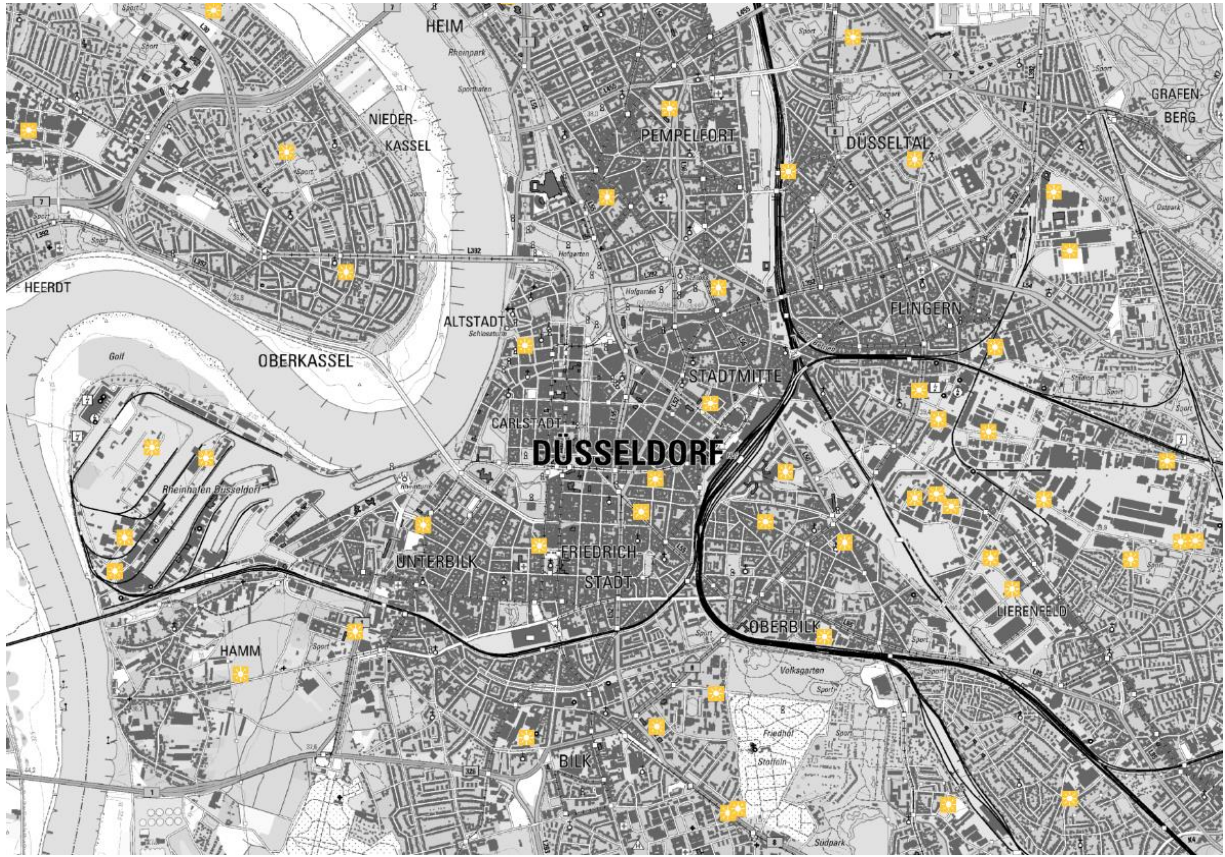

Figure 5: Energieatlas NRW, Herausgeber: Landesamt für Natur, Umwelt und Verbraucherschutz NRW

### 4.2.2. Building Polygons NRW

Many solar photovoltaic panels are small scale installations that can be found on the roofs of commercial structures, or private homes. In order to extract these solar panels, building polygons can be used to concentrate the search at areas where buildings are located. The Official Building Polygons of NRW are available as open data and are published by the NRW district government in Cologne. The polygons of the Shape files only contain geometric information and include no information about building usage, building age or any information about its owner.

The figure underneath shows an example of the building polygons and the combination of building polygons with DOP20:

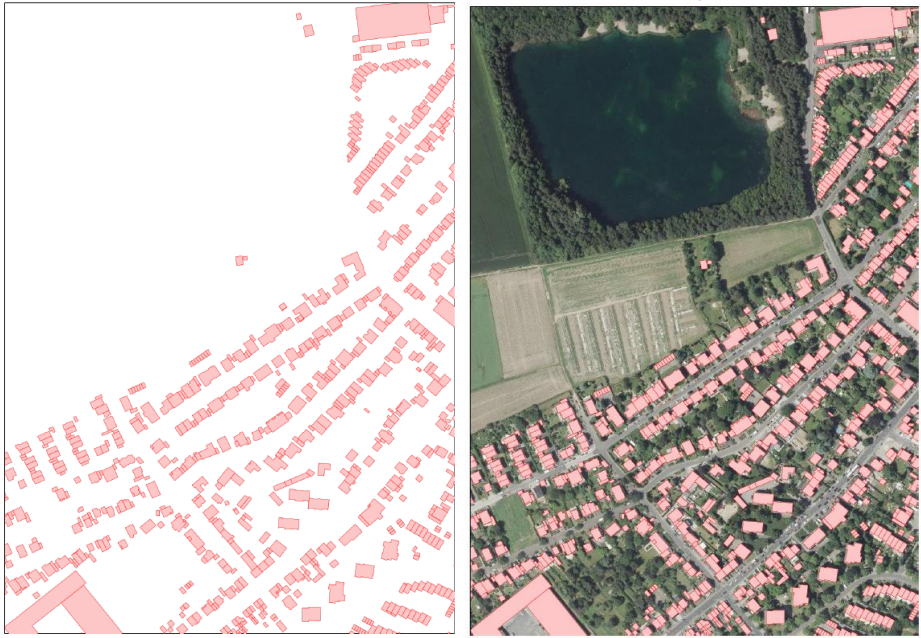

Figure 6: Geobasisdaten der Kommunen und des Landes NRW © Geobasis NRW. No official standard version

# 5. Belgium – Flanders

In Belgium territorial competencies are exercised not at the Belgian federal level but at the level of its three regions: Flanders, Wallonia and Brussels. Consequently, aerial photography and auxiliary administrative datasets need to be collected, if available, from three separate administrations functioning totally autonomously. Therefore the overview below is presented by region.

## 5.1. Aerial photography

The Agentschap voor Geografische Informatie Vlaanderen (AGIV) annually commissions the creation of an area-wide medium-scale orthophoto coverage of the Flemish Region, including the Brussels-Capital Region. This assignment is in two phases: first, the realization of digital photographic aerial photographs in the winter flight season with a ground resolution of 17cm and, secondly, the production of an orthophotomosaic with a ground resolution of 25cm. This product is a compilation of the orthophotomosaics (winter photographs) that were acquired for each part of this area in 2016. The product is supplemented with a data set 'flight day contour' that shows the date of recording for each part of the compilation. These Aerial photo datasets are available via the **EODaS (*Earth Observation Data Science*) unit** of the Department 'Omgeving' ('Environment'). Also available are

- Three-yearly summer recordings @40cm;
- 10-yearly (in principle) LiDAR recordings (LiDAR@25cm, RGB @10cm), most recent ones 2013-2015.

The datasets are freely available after registration on the website of This page (https://overheid.vlaanderen.be/informatie-vlaanderen/producten-diensten/earth-observation-data-science, in Dutch) provides an overview of the available remote sensing data and the conditions under which they can be obtained (open and free of charge, in principle).

The user can browse through the data thank to the AGIV BVK viewer, here

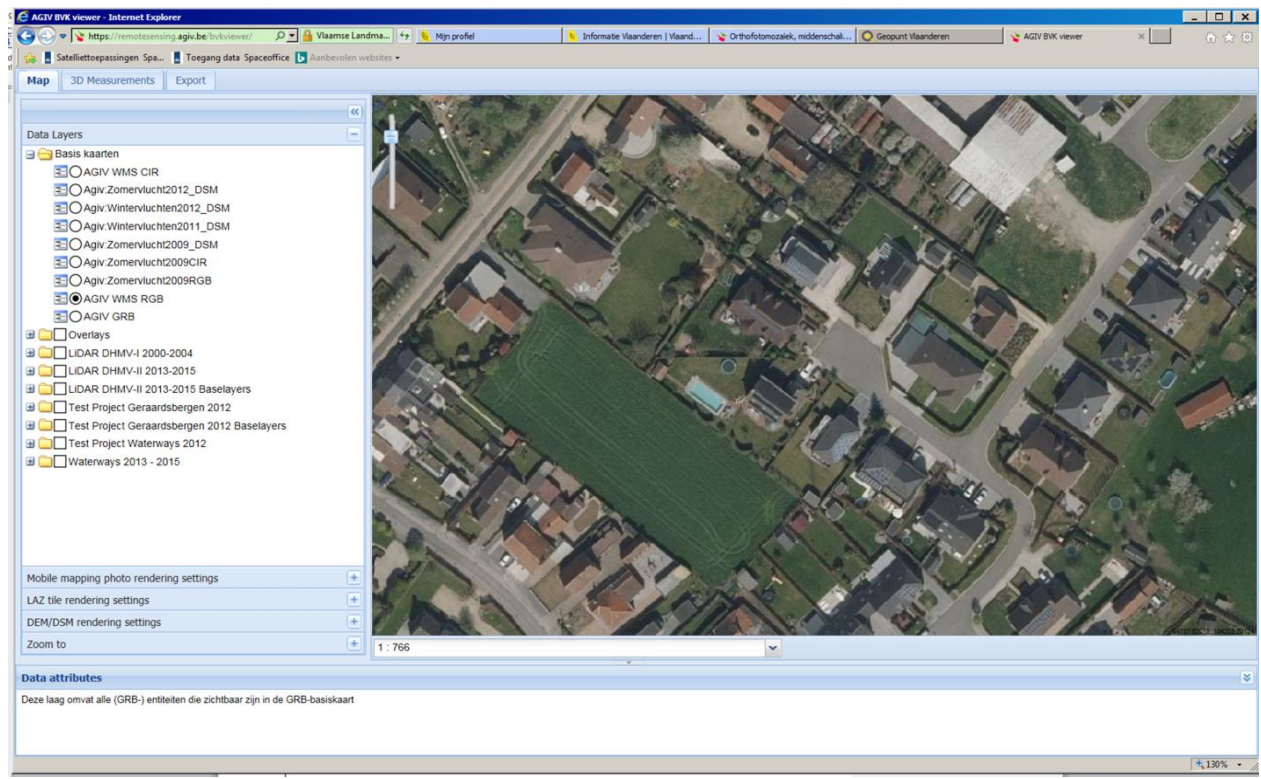

Figure 7: Snapshot of the AGIV BVK viewer.

**Contact has been made with EODaS on behalf of this project and collaboration in principle agreed.**

## 5.2. Administrative Datasets

175000 sets of panels have been installed in Flanders, but it's thought that one in six owners has failed to register them with the network providers[5]
In the context of the "prosumer tariff" (a sum of about 100€ per kilowatt solar panel owners have to pay to send electricity to the network) the network providers Eandis and Infrax (soon to be merged to Fluvius) should know perfectly who has installed how many solar panels (see https://www.vlaanderen.be/nl/bouwen-wonen-en-energie/elektriciteit-aardgas-en-verwarming/prosumententarief-voor-eigenaars-van-zonnepanelen-windmolens-en-wkk-installaties-kleine-installaties  andhttp://docs.vlaamsparlement.be/pfile?id=1337970, in Dutch).

---

[5] http://www.flanderstoday.eu/current-affairs/solar-panel-owners-must-register-or-face-fine

These data are not public but are available for the administration and statistics (for instance for statistics at municipal level, see https://www.energiesparen.be/cijfers/zonnepanelen, in Dutch).

Furthermore, in order to estimate the potential, EODaS has developed for the Vlaams Energieagentschap(VEA) the 'sunshine map'[6] (). For this LiDAR was used, allowing to estimate the potential of each building. A short movie illustrating this map can be found on Youtube[7].

# 6. Satellites Datasets

One of the research questions of this project is to determine the required minimum resolution for detecting solar panels. To answer this question, satellite images with different spatial resolutions will be tested: A satellite image with a spatial resolution below 1 m will be used, which is the highest resolution available and is therefore the most promising. Furthermore, images with a resolution of 1-2 m, 5 m and 10 m will be used. The results of the higher resolutions will tell, whether using resolutions as low as 5 or 10 m make sense for the detection of solar panels.

There are three main requirements for the satellite images:

- The cloud coverage of the satellite images should be minimal. A meaningful analysis is only possible when no clouds or cloud shadows obscure the image.
- The recording date of the image should be no later than December 2017 to match the currency of the administrative data of LANUV. Ideally, the images should be from the years 2016 to 2017 to match the aerial images.
- Furthermore, the satellite images should at least roughly match the test area used in the aerial images. If this is not possible, the study area should be heterogeneous regarding rural and urban areas.

**Error! Reference source not found.** shows the average cloud cover over Europe during the months January, April, July and October 2015. Some seasonal variation can be seen but what is most striking is that parts of northern Europe are extremely clouded throughout the year. On average 55 % of the land is covered by clouds with seasonal and spatial variation ([KPS+13]).

---

[6] Available at https://www.energiesparen.be/zonnekaart
[7] Sunshine map movie:https://www.youtube.com/watch?v=NWHEAxvBeQU

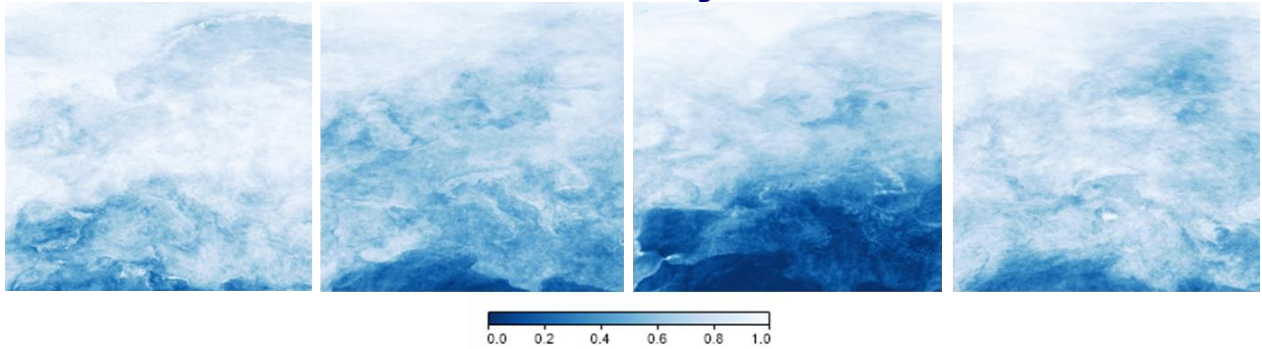

Figure 8 :Average cloud cover in 2015 across Europe for January, April, July and October (from left to right) based on TERRA/MODIS observation. (Source: Nasa Earth observationi:)

An ad hoc analysis of the study area was carried out to determine the average cloud coverage. Sentinel images can be found at the Copernicus Science Hub (https://scihub.copernicus.eu/). The metadata of these images, such as cloud coverage, are published alongside. We analysed the cloud coverage of all Sentinel-2 images from 2016 and 2017, which cover at least parts of the study area. Only 8.2 % of the images in this period had a cloud coverage lower than 5 %. A further 1.3 % of the images had a cloud coverage between 5 and 15 %. However, the given cloud coverage is only provided for the entire acquired satellite image which is significantly larger than the study area. Therefore, these numbers only provide an indication, that satellite images, which can be used for analysis, are scarce due to cloud coverage.

## 6.1. Datasets

The algorithm is used on satellite imagery with different resolutions: The higher the resolution, the more promising its results. What is of interest is to determine the required minimum resolution for the detection of solar panels. Four different resolutions might be investigated: below 1 m, 1-2 m, 5 m and 10 m. The exact resolution depends on the available satellite data, which fulfil the mentioned conditions, and the results of the use of a first set of high-resolution satellite data.

### 6.1.1. Below 1 meter resolution

The only very high-resolution satellite image fulfilling all of the aforementioned requirements are images taken on December $20^{th}$, 2016 by a Pleiades Satellite. Figure 9 shows the area, which is covered by the satellite image and which corresponds roughly with the study area.

The twin satellites Pleiades 1A and 1B were launched in 2011 and 2012, respectively. They are very high-resolution optical imaging satellites with a revisit time of 2 days. The multispectral images have a resolution of 2.5 m, which are resampled to 2 m at ground level. The panchromatic (black and white) images have a resolution of 70 cm, resampled to 50 cm. Through pan-sharpening (panchromatic sharpening) they are fused to achieve a multispectral image with a resolution of 50 cm.

Airbus Defence and Space have exclusive access to the Pleiades images. Destatis is buying the data from its only German distributor GAF AG. The study area of 450 km² covers a very heterogeneous area regarding rural and urban areas and with it various housing structures. The cost of acquiring this archive data is 5355 EUR. The high cost of very high resolution satellite imagery and low frequency of cloudless images have to be considered when answering the research question of whether it is possible to develop a harmonized method across EU member states.

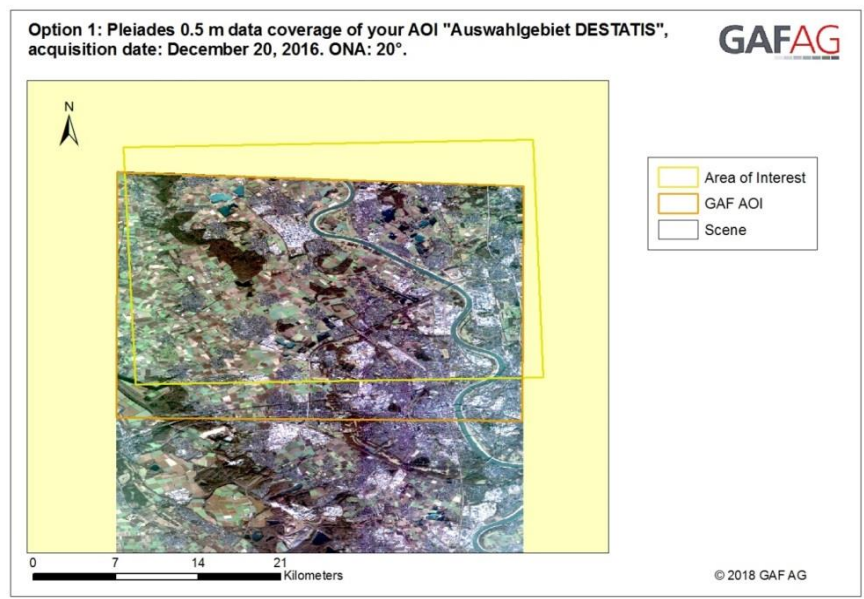

Figure 9: The yellow rectangle shows the area of interest (AOI) for this study. The orange rectangle refers to the AOI of the satellite image provider. The bought image is the overlay area of both AOIs. The entire acquired satellite image ("scene") is shown in grey (GAF AG)

The satellite image was taken at an off-nadir angle of 20 degrees, which means that the original image was optically slightly distorted by the angle. However, the image was orthorectified, which is the process of rectifying the distortion. 16 bit color was chosen over 8 bit color, which is more useful for image detection but has the disadvantage of a larger file size. Further satellites exist which have a similar resolution. For instance, images by the Ikonos satellite have a resolution of 82 cm. The satellite QuickBird has images with a resolution of 65 cm. Images with these high resolutions are commercial and have to be purchased. Which exact images will be used, still has to be decided.

### 6.1.2. 1-2 meters resolution

For the resolution of 1-2 meters the satellites Spot 6 and 7 could for example be used, which has a resolution of 1.5 m. However, alternatively the higher resolution images could be resampled to various resolutions. The advantages of this procedure are twofold: Images of this resolution are expensive to obtain and furthermore a lot of different resolutions can be tested.

### 6.1.3. 5 meters resolution

Destatis has access to RapidEye images through its cooperation with the federal agency for cartography and geodesy. However, they can only be used internally and cannot be shared with the project partners or published. RapidEye is a constellation of 5 satellites, owned and operated by the enterprise Planet Labs. Its images have a resolution of 6.5 m and 5 m orthorectified. RapidEye revisits every 5.5 days and off-nadir on a daily basis. The date and exact location of the images still have to be decided, but the choice will be based on the aforementioned requirements.

### 6.1.4. 10 meters resolution

Furthermore, the Sentinel-2 images will be used to test the performance of object detection with less resolution. These images have a resolution of 10 m, however some bands have an even lower resolution of 20 or 60 m. Sentinel-2 also consists of two satellites through which the earth can be observed every 5 days. The main advantage of using Sentinel-2 is its free and open data policy. Unlike some commercial satellites, the images do not have to be tasked to be taken, but are constantly being taken and stored. Sentinel-2 images have 13 bands. The date of the images and study area, still have to be decided. However, depending on the results of the higher resolution images it also has to be decided whether using this relatively coarse resolution is meaningful for household detection purposes. However, we do expect to be able to use these coarse resolution to identify larger solar plant.

## 6.2. Dutch satellite data portal

The access to some commercial satellite data will be facilitate in the case of the Netherlands as The Netherlands Space Office (NSO) has set-up a satellite data portal which provides access to satellite data from the Netherlands to Dutch users. This portal consists of several facilities. For example, there is a portal that provides (free) access to pre-processed satellite data from the Netherlands: a generic facility so that (high-resolution optical) satellite images and web services which allow for the download of data after registration/login by Dutch users. Besides, there is also (free) access to raw satellite data from the Netherlands, both optical and radar data. These are not ready-made products, but will first have to be pre-processed (e.g. geometric correction and orthorectification). These raw optical data and radar data can be downloaded via FTP (registration required). Figure 10 provides with a snapshot of the data portal provided by the Netherland Space Office.

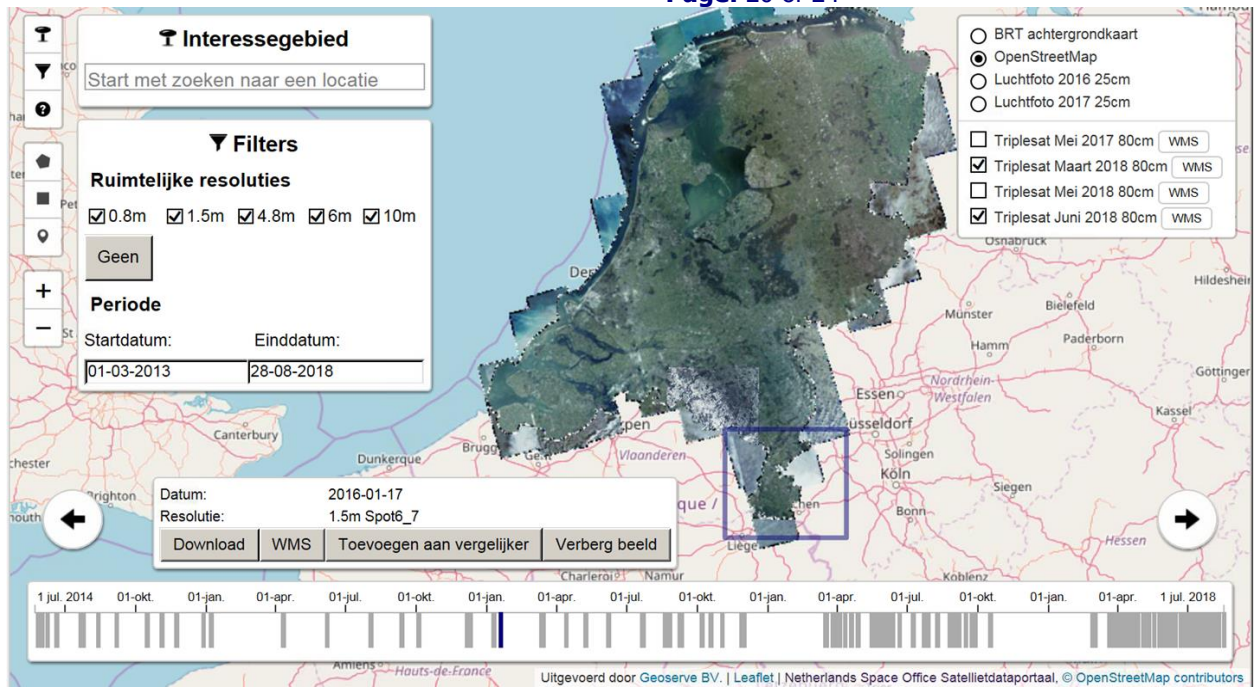

Figure 10: Snapshot of the data portal provided by the Netherland Space Office.

The Satellite Data Portal provides access to raw and pre-processed satellite data from the Netherlands, originating from numerous instruments on various satellites. The instruments measure actively as well as passively and with a variety of frequencies and resolutions. The following is an overview of the current offer of satellite data in the Satellite Data Portal, both archive data and recent data.

## 7. Software developed to process the aerial image

Aerial images have to be pre-processed in order to be analysed by machine learning algorithms. Software was written to create small tiles for a certain region of the total image. The tile size can be configured to be of an appropriate size as input for a machine learning algorithm.

Specifically, in order to download the tiles, we made use of OWSLIB or GDAL, that are Python and C++ libraries that allow to deal with the WMS protocol, a REST protocol to download geoTIFF images from HTTP services published online.

A version of the script used to download tiles in Python for Deep Solaris is available in [8], whereas an example of the C++ software is available in [9].

First of all there are a set of WMS services that allow us to simply query an HTTP link and get a map from an area, provided that we know the gps coordinates of an area. The most prominent WMS service from the Netherlands is PDOK and it allows to query with a 25cm per pixel and it allows for both RGB and Infrared

---

[8] https://github.com/SB-BISS/DeepSolaris/blob/master/Python/DeepSolarisAutomaticNRWAnalysis.ipynb
[9] https://github.com/thinkpractice/MapTiler

The link for RGB:
https://geodata.nationaalgeoregister.nl/luchtfoto/rgb/wms?&request=GetCapabilities
The link for Infrared:
https://geodata.nationaalgeoregister.nl/luchtfoto/infrarood/wms?&request=GetCapabilities

The Netherlands register allows also to query by year, with a maximum 5 years period, which may allow reasoning of multiple types concerning the map. The link for NRW (North Rhine Westphalia) provides both RGB and Infrared possibilities in the layers that can be directly queried in the http service https://www.wms.nrw.de/geobasis/wms_nw_dop

Concerning the Flanders region, in addition to the aforementioned possibilities to download the data, a WMS link is also provided:
https://remotesensing.agiv.be/gis/geoserver/wms?REQUEST=GetCapabilities

The written software, has a configurable pipeline in which several aerial images can be processed at once. In a standard application, first the tiles are generated for a certain bounding box of a geographical region. Next, the images for each tile are retrieved from aerial images for different years and spectrums. After that, polygon and address metadata are added for each tile and saved in a database. The building polygons are then used to create mask images with which the original tile is masked. In this way, several images for a certain tile can be written at once (see figure 3). The software was written to process large amounts of image data at once and can handle a wealth of geographical image formats. It will be used to generate large sets of tiles to be used with the machine learning algorithms. The annotation of these tiles with register data will simplify the generation of training sets for machine learning.

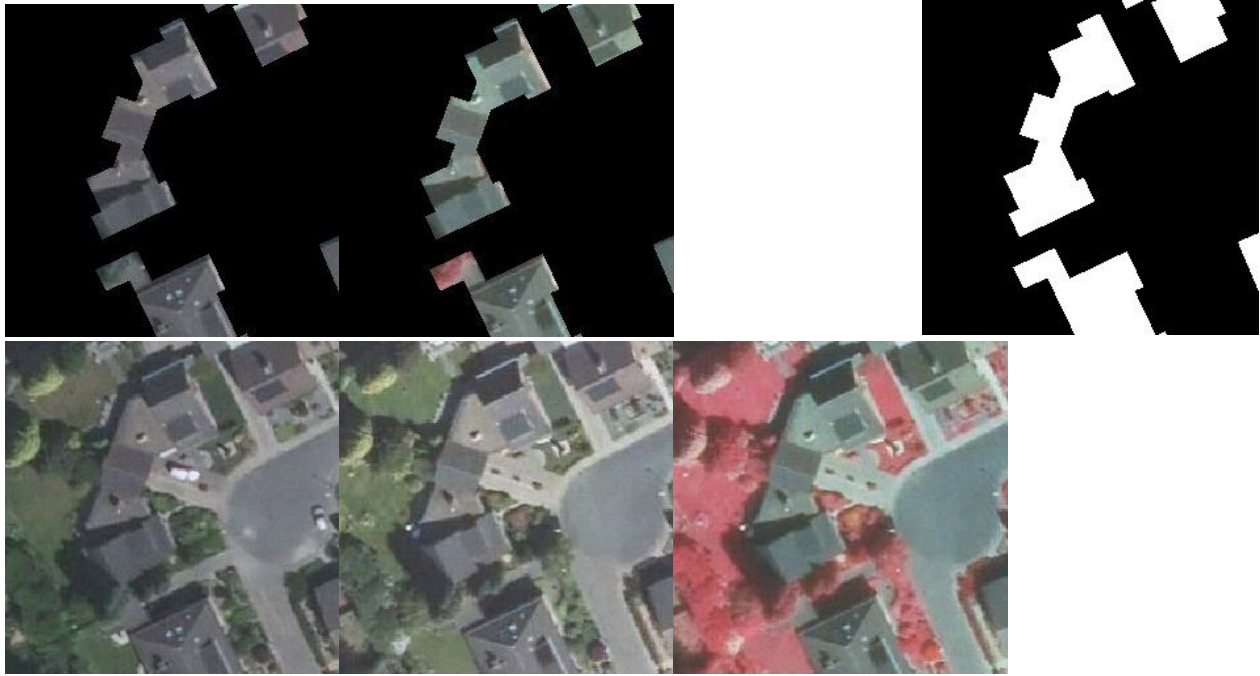

**Figure 11: Several images generated for one geographical tile; in clockwise fashion, masked images, mask, ir 2016, rgb 2016, rgb 2017.**

# 8. Conclusion

This report discusses the available remote sensing data in the regions of NRW, Netherlands and Flanders. Several possibilities to access the data are available, including direct download or download by means of REST services. The three regions all provide REST services for the analysis of GIS data. The resolution of the images in the Netherlands region is limited to 25cm per pixel, the resolution for the NRW reaches a max resolution of 10cm per pixel, whereas Flanders present a similar data set to that available in the Netherlands.

In addition to the GIS data of the region, additional administrative data sets can be used in combination with the GIS data. Such data set can include information concerning the buildings in the image, but also the presence of registered solar panels (as in the case of LANUV data).

Finally, satellite images can be used to obtain additional information concerning the insulation in a region during a certain period of the year.

Future work will build on the inventory of data sets defined in this contribution in order to identify solar panels in the images, for object detection in a first instance, with a binary classification of the images, and for object recognition in a second instance, to identify the pixels that represent the solar panel in the image.

---

[i] **https://neo.sci.gsfc.nasa.gov/view.php?datasetId=MODAL2_M_CLD_FR&date=2015-04-01** last access on Augst 2018